\documentclass[12pt]{article}

\usepackage{graphicx}
\usepackage{amsmath,amssymb,color,times,float,amsthm}
\usepackage{subfig}
\usepackage{mathptmx}      
\newtheorem{definition}{Definition}
\newtheorem{theorem}{Theorem}
\newtheorem{corollary}{Corollary}[theorem]

\begin{document}

\pagestyle{empty}
\begin{center}
{\bf \large Understanding Adversarial Examples Through Deep Neural
  Network's \\[0.15in] Response Surface and Uncertainty Regions} \\[0.15in]
Juan Shu\\
Department of Statistics, Purdue University\\
Email:~shu30@purdue.edu\\[0.1in]
Bowei Xi \\
Department of Statistics, Purdue University\\
Email:~xbw@purdue.edu \\[0.1in]
Charles Kamhoua \\
US Army Research Laboratory \\
Email:~charles.a.kamhoua.civ@mail.mil
\end{center}

\paragraph{Abstract:}
Deep neural network (DNN) is a popular model implemented in many  
systems to handle complex tasks such as image classification, object
recognition, natural language processing etc. 
Consequently DNN structural vulnerabilities become part of the security
vulnerabilities in those systems. 
In this paper we study the root cause of DNN adversarial examples. 
We examine the DNN response surface
to understand its classification boundary. Our study reveals the
structural problem of DNN classification boundary that leads to the adversarial
examples. Existing attack algorithms can generate from a handful
to a few hundred adversarial examples given one clean image. We show
there are infinitely many adversarial images given one clean
sample, all within a small neighborhood of the clean
sample. We then define DNN uncertainty regions and show
transferability of adversarial examples is not universal. We also argue that
generalization error, the large sample 
theoretical guarantee established for DNN, cannot adequately capture the phenomenon of
adversarial examples. We need new theory to measure DNN robustness.

\paragraph{Key Words:}
Adversarial Machine Learning, Deep Neural
Network, Response Surface, Uncertainty Regions, DNN Classification Boundary

\section{Introduction}
\label{sec:intro}

DNN is a powerful tool for  
complex tasks, such as image classification, object
recognition etc., especially when there are thousands of object
classes. Pictures of 
connected nodes are often displayed to illustrate 
the structure of DNN. DNN generates feature maps from an input that
are most strongly associated 
with the learning task. Pooling layers, such as max pooling, can
further identify the features. And the popular activation function, the
Rectified Non-Linear Unit (ReLU), aims to solve the vanishing gradient
problem \cite{book-deep-learning-2016}. One earliest application of DNN
was to classify handwritten images \cite{lenet1998}. 
Starting with the success of
AlexNet on the ImageNet classification task \cite{alexnet-2012}, DNN has
regained popularity over other classifiers. Soon afterwards,
researchers noticed that 
targetedly adding minor perturbations to a clean image can cause a DNN
to misclassify the perturbed image \cite{attack-LBFGS}. This is the
beginning of a new chapter on 
adversarial machine learning research.

There are real world implications when using DNN in critical
applications without fully understanding its properties, such as its
classification boundary and vulnerabilities.  For example, 
Tesla uses camera and radar as sensors. DNN is used
to process the videos received from cameras, as described on the
webpage for Tesla AutopilotAI.
~\footnote{https://www.tesla.com/autopilotAI (last accessed Apr.18,
  2021)}
Soon Tesla Model 3 and Y vehicles will ditch radar and rely only on
cameras. 
~\footnote{https://www.cnbc.com/2021/05/25/tesla-ditching-radar-for-autopilot-in-model-3-model-y.html
  (last accessed June 14, 2021)}
In a system where the algorithm used to process sensor data has inherent
vulnerabilities, they become part of the security vulnerabilities the
attackers can explore.  
Notice Tesla suffered high profile
fatal crashes where its Autopilot was suspected to play a role.
~\footnote{https://abc7.com/tesla-fontana-investigation-fatal/10634134/
  (last accessed May.20, 2021)}
~\footnote{https://abcnews.go.com/Business/tesla-autopilot-mode-crashes-parked-police-car/story?id=77753735
  (last accessed May.20, 2021)}

The cause of the adversarial examples is a mystery until now. So far there
are only conflicting conjectures.  
\cite{attack-LBFGS,explain-adexample-2019} believed adversarial
examples lie in ``dense pockets'' in lower dimensional manifold, caused
by DNN's non-linearity. On the other hand
\cite{explain-adexample-2015} believed it is DNN's linear nature
and the very high dimensional inputs that lead to the adversarial
examples.
Furthermore \cite{explain-adexample-2015} used Figure 4 to show ``adversarial
examples occur in contiguous regions of the 1-D subspace defined by
the fast gradient sign method, not in fine pockets.'' 
Although Explainable Artificial Intelligence
(XAI) becomes a hot research area, aiming to decipher DNN internal
components, there is still a lack of
understanding of one most basic concept of DNN -- the shape of its
classification boundary.
For example, \cite{transferable-adexample-space-2017} showed
the classification boundary of DNN as a curve in Figure 3, and the
adversarial region was on the other side of the classification boundary in
Figure 1. 

In this paper we study DNN's classification boundary through its
response surface. Through experiments, we show the problem of
adversarial examples is not as simple as linear vs. non-linear. It is
a far more complex structural problem. 
Response surface methodology is a well studied subject in
statistics. A response variable $y$ is influenced by several
independent variables $x_1$, ..., $x_k$. The relationship is modeled
or approximated by a function with an error term,
$y=h(x_1,...,x_k)+\epsilon$. The response surface describes the
relationship between $y$ and $x_1$, ..., $x_k$. 
Response surface methodology plays an important role in
design of experiments, e.g.,  
\cite{response-surface-2010,book-response-surface-2016}. We notice
that along with the research on response surface methodology, 
design robustness and design uncertainty received considerable
attention since
\cite{box-draper-1959,box-draper-1963,box-draper-1975}.

We know the response surface of many well known models, such as
linear regression, generalized linear regression, non-parametric regression,
to name a few. For supervised learning techniques, support vector
machine (SVM) uses hyper-planes to separate the data points; linear
discriminant analysis has a linear decision boundary; quadratic
discriminant analysis has non-linear decision boundary. Despite many work on
building a robust DNN and to evaluate DNN robustness, we are yet to
know the shape of DNN classification boundary.
Without the knowledge of DNN classification boundary, 
  building a robust DNN model will remain an elusive task. 
The most significant contributions of this paper are the following.
\begin{enumerate}
\item We show DNN classification boundary is highly fractured, unlike other
  classifiers. The adversarial examples exist
  in lower dimensional hyper-rectangles within
  a small neighborhood surrounding each clean image,

\item We show that transferability of adversarial
  examples is not universal, contrary to 
  \cite{attack-LBFGS,explain-adexample-2015,transferable-adexample-space-2017}. 
  \cite{attack-LBFGS,explain-adexample-2015,transferable-adexample-space-2017}
  suggested that adversarial examples generated against one DNN are 
  misclassified by other DNNs, even if they have different model
  structures or are trained on different subsets of the training
  data. We show that adversarial examples against one DNN can be
  correctly classified by other DNN models, simply by using different
  initial random seeds in the training process. This leads to our
  definition of DNN uncertainty regions.        
 
\item We argue that generalization
  error, which measures DNN large sample performance, 
  cannot adequately capture the phenomenon of adversarial examples. 
  New theory is needed to measure DNN robustness. 
 
\end{enumerate}  

Besides the three major contributions, additional contributions of
this paper are the following. 
\begin{enumerate}
\item Given one clean image, existing attack algorithms generate up to 
  a few hundred adversarial examples through an optimization
  approach. Sampling from the lower dimensional hyper-rectangles lead
  to a stronger attack -- we have infinitely
  many adversarial examples given one clean image. 

\item Far fewer pixels are perturbed to form these
  hyper-rectangles compared to the existing attack
  algorithms. Therefore we reduce the total
  amount of perturbations added to a clean image to create
  adversarial examples.    

\item Training a DNN is a non-convex optimization problem. There are
  many local optima. By using different initial values, we obtain
  different trained DNN models. Some benchmark attacks, such as
  Carlini \& Wagner $L_2$ attack, often attacks the target
  model only. Our adversarial examples can attack 
  multiple models simultaneously.
  
\end{enumerate}  

The paper is organized as follows. Section~\ref{sec:related} discusses 
the related work. Section~\ref{sec:uncertain} conducts experiments to show
DNN response surface and introduces the concept of DNN uncertainty
regions. Section~\ref{sec:theory} discusses the 
discrepancy between the theoretically proven DNN generalization error bound and the existence
of adversarial examples. 
Section~\ref{sec:conclusion} concludes this paper. 

\subsection{Related Work}
\label{sec:related}

There is an extensive literature on robust statistics, e.g.,
\cite{robust-2004,robust-2000}.
Robust regression, robust tests, and robust estimators have been developed to address
the problems where the true distribution deviates from the
model assumption, and outliers. Adversarial machine learning
also aims to build robust learning models, but under a  
different scenario -- adversaries discover the vulnerabilities of a
learning model, which cause the learning
model to make a costly mistake. There
are two broad categories of attacks, poisoning attacks and evasion
attacks \cite{xi-wires-2020}. Poisoning attacks inject malicious
samples into the training data, to cause the resulting learning model
to make a mistake with certain test samples. Assuming there is no easy
access to the training process, evasion attacks generate test samples
that the learning model cannot handle correctly. The adversarial
examples generated to attack DNN belong to evasion attacks.   

Adversarial (evasion) attacks against DNN are the earliest
attacks. Recently there are attacks designed to 
break graph neural network (e.g., \cite{attack-gnn-1,attack-gnn-2}),
recurrent neural network (e.g.,
\cite{attack-rnn-1,attack-rnn-2,attack-rnn-3,attack-rnn-4,attack-rnn-5}),
and deep reinforcement learning (e.g.,
\cite{attack-RL-1,attack-RL-2,attack-RL-3,attack-RL-4}).   
In this paper, we study the response surface and uncertainty
regions of DNN, i.e., convolutional neural network (CNN) or fully connected
neural network (multilayer perceptron (MLP)).
There are many existing attacks against such DNN
models. In our experiments we use Foolbox \cite{foolbox}, which implemented 42 attack
algorithms. Depending on adversaries' knowledge of a DNN model, there
are white-box attacks and black-box attacks. For white-box attacks,
adversaries know the true DNN model, including model structure and
parameter values. For black-box attacks, adversaries don't know the
true model. Instead, adversaries query the true model, build a
local substitute model based on the queries, and attack the local
model. A targeted attack generates adversarial examples
that are misclassified into a pre-determined class, while an
untargeted attack simply generates misclassified samples.  

Several survey papers are published, introducing the
current state and the timeline of attacks and defenses, e.g.,
\cite{survey-1,survey-2,xi-wires-2020}. In
general, the attack algorithms follow an optimization approach, i.e.,
generating adversarial examples while minimizing their distances to
the clean samples.
Let $W$ be a clean image and $W^a$ be an adversarial
example. Let $M$ be a trained DNN model that assigns a class label to
$W$. We have $M(W^a)\neq M(W)$. 
$W$ is a matrix for a gray-scale image, and a tensor for a color
image. The size of the matrix/tensor is determined by the image
resolution. The individual elements (pixels) in $W$ represent the
light level, having integer values ranging from 0 (no light) to 255
(maximum light). The pixels are often rescaled to $[0,1]$. $W$ can be
vectorized. Our approach needs close to a hundred adversarial images
$W^a$ given one clean image $W$.  Some attack algorithms generate 
a single $W^a$ or only a handful of $W^a$s are not used in our
experiments. We also exclude attack algorithms that need large
perturbations to generate $W^a$, since the resulting adversarial
images will be outside of a small $\delta$ neighborhood of the
clean image $W$.  Here we introduce the attack algorithms
that are used in our experiments. 
\begin{itemize}
\item Pointwise (PW) Attack \cite{foolbox-2017}: An effective
  decision-based attack minimizing $||W^a-W||_0$, i.e., the number of
  pixels in $W$ and $W^a$ that have different values.    

\item Carlini \& Wagner $L_2$ (CW2) Attack \cite{CW2-attack-2017}:
  $W^a$ is generated by minimizing $\textrm{Distance}(W,W^a) + a 
  \times \textrm{loss}(W^a)$. $L_1$, $L_2$, and $L_{\infty}$ norms are
  considered. The $L_2$ distance is the most commonly used.   

\item NewtonFool (NF) Attack \cite{NewtonFool-attack-2017}: The
  perturbation follows the gradient that reduces the probability $W^a$
  belongs to the correct class $c$.

\item Fast Gradient Sign Method (FGSM) \cite{explain-adexample-2015}:
  Given a loss function,   
  $W^a$ is generated as $W^a = W - \epsilon\times 
  \textrm{sign}(\nabla \textrm{loss}(W))$. $\epsilon$ is the
  step size, usually a small number for producing minor
  perturbation. FGSM follows the sign of the gradient of the loss
  function. It generates adversarial examples much faster than Carlini
  \& Wagner attack.    

\item Basic Iterative Method (BIM)  \cite{BIM-attack-2017}: It
  improves FGSM attack by clipping the pixel values in each iteration
  to reduce the amount of perturbation added to $W$. $L_1$, $L_2$ and
  $L_{\infty}$ distance can be used in the attack. 
 
\item Moment Iterative (MI) Attack
  \cite{moment-iterative-attack-2018}: It applies momentum to
  accelerate the gradient descent in a set-up similar to FGSM while
  escaping local maxima with poor results.  
  
\end{itemize}

\section{DNN Response Surface and Uncertainty Regions}
\label{sec:uncertain}

DNN's response surface is described by $M(W)=c$, where $c$ is the
object class assigned to image $W$ by a trained DNN model $M$. In this
paper we assume $M$ assigns hard labels. The
response surface is more complex, when $W$ is a high
resolution color image with $M$ 
assigning soft labels, and classification accuracy is assessed using top 5
classes. We leave it to the future work. 

Uncertainty regions are proposed for SVM facing multi-class
classification task. For one-against-all SVM, multiple separating hyper-planes are used to
classify the samples. The areas within the margins of the binary 
hyper-planes are the SVM uncertainty regions 
\cite{svm-uncertain1,svm-uncertain2,svm-uncertain3,svm-uncertain4}.
Meaning if a data point is very close to multiple binary decision
boundaries, SVM is uncertain which class it belongs to.  

We propose a different definition of DNN uncertainty regions.
Given a training dataset $D_n$ with $n$ samples, let $\mathcal{M}=\{M_1, M_2, ...\}$ be
the set of DNN models with
identical model structure, i.e., same number of layers, same activation,
etc. $M_i \in \mathcal{M}$ is obtained by having a DNN model trained on
$D_n$ and varying the initial seeds. Since DNN 
training is a non-convex optimization problem, they converge to
different local optima. For $i \neq j$, $M_i$ and $M_j$ have
different parameter values. Define DNN uncertainty region as follows.  
\begin{definition}
An uncertainty region is defined as
$U~:=~\left\{W: \exists~ M_i,~M_j ~\in \mathcal{M},~ s.t.~M_i(W)~\neq ~M_j(W)\right\}$. 
\end{definition}
In a small $\delta-$neighborhood of a clean image $W$, $\delta >0$, 
a perturbed image $W^a$ is a noisy but
recognizable version of the clean image $W$. We use the $L_2$ distance
between $W$ and $W^a$, $d(W,W^a)=||W -W^a||_2$. Define $B(\delta,W)$ as 
$B(\delta,W)~:=~\left\{W:~ d(W^a,W) \leq \delta \right\}. $   
Inside $B(\delta,W)$ with $\delta$ sufficiently small, ideally a classifier
should assign all the noisy images to the same object class of $W$. 

\begin{figure}[tb]
\centering
\includegraphics[width=5in]{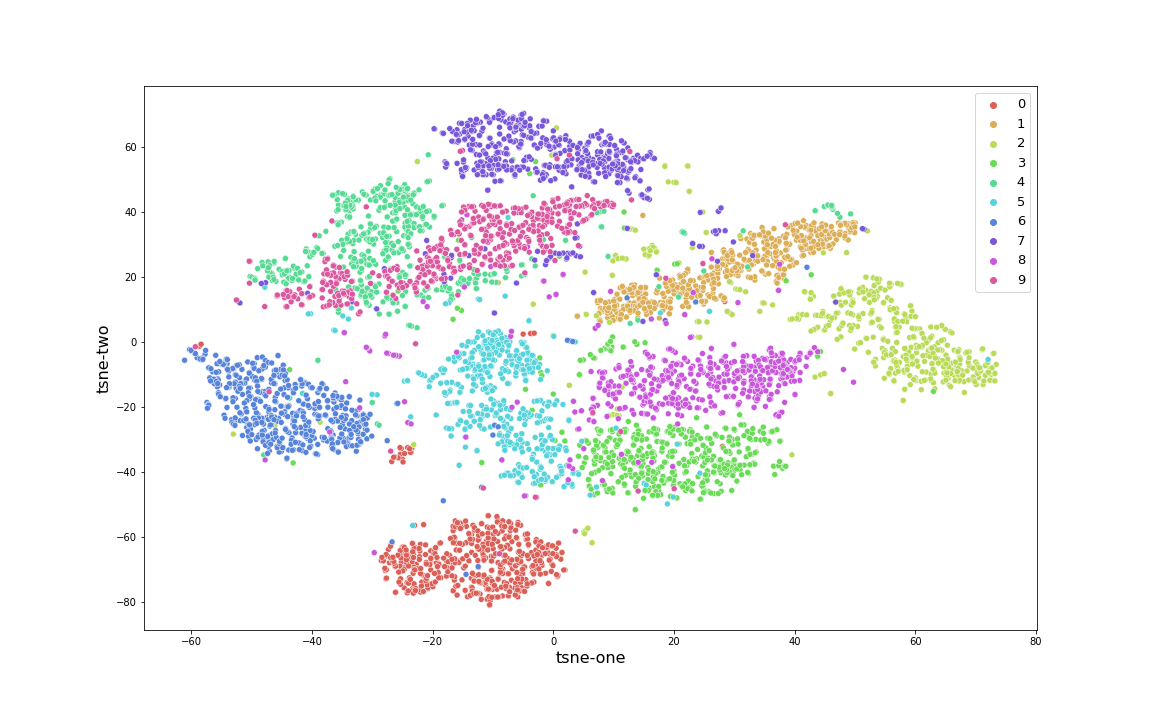}
\caption{Lower Dimensional Projection of MNIST data}
\label{fig:2D}       
\end{figure}
\begin{table}[tb]
\centering
\caption{CNN Mis-classification Rates on Clean MNIST Test Images}
\label{tab:baseline}
\begin{center}
{\footnotesize  
\begin{tabular}{|llllllllll|}
\hline
$M_1$&$M_2$&$M_3$&$M_4$&$M_5$&$M_6$&$M_7$&$M_8$&$M_9$&$M_{10}$\\\hline 
0.033 & 0.035 & 0.025 & 0.019 & 0.017 & 0.015 & 0.013 & 0.012 & 0.012 & 0.012 \\
\hline 
\end{tabular} }
\end{center}
\end{table}

\begin{figure}[tb]
\centering
\includegraphics[width=0.25\textwidth]{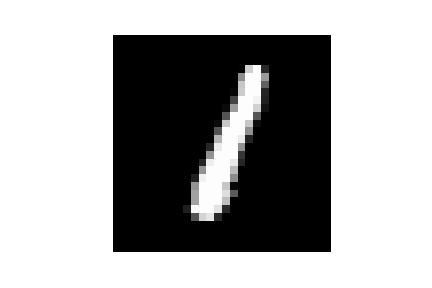}
\caption{Clean Test Image 1}
\label{fig-1}
\end{figure}

\begin{table}[t]
\caption{CNN Misclassification Rates in Hyper-Rectangles}
\label{tab:hyper-rec-misrate}
\begin{center}
{\footnotesize  
\begin{tabular}{|l|l|llllllllll|} \hline
  &$s_{(i)}$ &$M_1$&$M_2$&$M_3$&$M_4$&$M_5$&$M_6$&$M_7$&$M_8$&$M_9$&$M_{10}$\\\hline
PW 30d $1\to 2$ & 1 & 0.843&0.454&0.281&0.344&0.183&0.58&0.687&0.596&0.709&	0.848 \\   
CW2 280d $1\to 2$& 0.0001 & {\bf 0.929} &  {\bf 0} & {\bf 0} & {\bf 0}
& {\bf 0} & {\bf 0} & {\bf 0} & {\bf 0} & {\bf 0} & {\bf 0} \\
BIM $L_1$ 240d $1\to 2$&0.016 & 0.429 & 0.445 & 0.461 & 0.427 & 0.406 & 0.426 & 0.453& 0.441 & 0.444 & 0.450 \\
NF 60d  $1\to 2$& 0.030 & {\bf 0} & {\bf 0} & {\bf 0} & {\bf 0.001} &
{\bf 0.004} & {\bf 0.925} & {\bf 1} & {\bf 0.58} & {\bf 0.991} & {\bf 1} \\
FGSM 375d $1\to 2$& 0.012 & {\bf 0.915} & {\bf 0} & {\bf 0} & {\bf 0}
& {\bf 0} & {\bf 0} & {\bf 0}& {\bf 0} & {\bf 0} & {\bf 0} \\\hline
CW2 210d $1\to 3$&0.008 & 0.855 & 0.01 & 0.843 & 0.974 & 1 & 1 & 0.998 & 0.984 & 0.999& 1  \\  
BIM $L_1$ 230d $1\to 3$&0.016 & 0.458 & 0.456 & 0.484 & 0.482 & 0.472 & 0.466 & 0.462 & 0.469 & 0.497 & 0.480 \\
BIM $L_2$ 200d $1\to 3$&0.019 & 0.706 & 0.734 & 0.669 & 0.698 & 0.698 & 0.672 & 0.701 & 0.718 & 0.685 & 0.708 \\
MI 320d $1\to 3$&0.036 & 0.91 & 0 & 0 & 0 & 0 & 0 & 0 & 0 & 0 & 0 \\\hline
CW2 180d $1\to 4$ &0.009 & 0.999 & 0 & 0 & 0 & 0 & 0 & 0 & 0 & 0 & 0\\\hline
CW2 175d $1\to 5$ &0.011 & 0.047 & 0     & 0.135 & 1 & 1 & 1 & 1 & 1 & 1 & 1 \\
MI 205d $1\to 5$&0.036 & 0.991 & 0.041 & 0 & 1 & 0.916 & 1 & 0.845 & 0.035 & 0.907 & 1 \\\hline
CW2 150d $1\to 6$&0.013 &  0 & 0.099 & 0 &  0.993 & 0.007 &
0.945 & 0.807 & 0.042 & 1 & 0.652 \\
BIM $L_2$ 220d $1\to 6$&0.003 & 0.448 & 0.455 & 0.472 & 0.454 & 0.407 & 0.432 & 0.421& 0.449 & 0.431 & 0.421 \\\hline
CW2 220d $1\to 7$&0.003 & 0.871 & 0 & 0 & 0 & 0 & 0 & 0 & 0 & 0 & 0 \\
BIM $L_1$ 200d $1\to 7$&0.028 & {\bf 0.826} & {\bf 0.81}  &  {\bf
  0.824} & {\bf 0.812} & {\bf 0.832} & {\bf 0.811} & {\bf 0.825} &
{\bf 0.813} & {\bf 0.801} & {\bf 0.828}\\
BIM $L_{\infty}$ 380d $1\to 7$&0.002 & 0.782 & 0 & 0 & 0 & 0 & 0 & 0 & 0 & 0 & 0\\
MI 300d $1\to 7$&0.028 & 0.999 & 0 & 0 & 0 & 0 & 0 & 0 & 0 & 0 & 0\\\hline
PW 35d $1\to 8$&1 &0.809&0.818&0.48&0.856&0.896&0.974&0.932&0.754&0.92&0.937 \\
CW2 190d $1\to 8$&0.0004 & 0.847 & 0 & 0 & 0.014 & 0.655 & 0.009 & 0 & 0 & 0 & 0 \\
BIM $L_1$ 240d $1\to 8$&0.002 & 0.389 & 0.37  & 0.365 & 0.363 & 0.365 & 0.37  & 0.365 & 0.384 & 0.36  & 0.351\\
MI 280d $1\to 8$&0.006 & 0.908 & 0.909 & 0.908 & 0.900 & 0.930 & 0.919 & 0.904 & 0.911 & 0.918 & 0.917\\\hline
CW2 200d $1\to 9$&0.025 & 0.995 & 0 & 0.01 & 0.97 & 1 & 1 & 0 & 0.022 & 0.396 & 0.99\\
BIM $L_1$ 140d $1\to 9$&0.013 & 0.651 & 0.001 & 0.008 & 0.252 & 0.503  & 0.758 & 0.071 & 0.244 & 0.013 & 0.182\\\hline
CW2 200d $1\to 0$ & 0.005& {\bf 0} & {\bf 0} & {\bf 0} & {\bf 0} &
{\bf 0}     & {\bf 1} & {\bf 0.946} & {\bf 0} & {\bf 1} & {\bf 1}\\
BIM $L_1$ 120d $1\to 0$&0.036 & 0.761 & 0.714 & 0.751 & 0.724 & 0.738 & 0.748 & 0.722& 0.742 & 0.743 & 0.752\\
BIM $L_2$ 120d $1\to 0$&0.030 & 0.928 & 0.95  & 0.935 & 0.927 & 0.924 & 0.931 & 0.937 & 0.93  & 0.92  & 0.932\\
BIM $L_{\infty}$ 350d $1\to 0$&0.003 & 0.527 & 0.868 & 0.127 & 0.227 & 0.02 & 0.007 &0 & 0 & 0 & 0 \\
MI 230d $1\to 0$&0.032 & 0.996 & 1 & 0 & 0.235 & 0 & 0 & 0 & 0 & 0 & 1\\
\hline
\end{tabular} }
\end{center}
\end{table}

\begin{table}[t]
\caption{$L_2$ Distance}
\label{tab:L2}
\begin{center}
{\footnotesize  
\begin{tabular}{|l|lll|lll|} \hline
  &$L_2^{min}$ & $L_2^{max}$  &  $\bar{L}_2$ 
  &$L_2^{min}$ Attack& $L_2^{max}$ Attack & $\bar{L}_2$ Attack\\\hline
PW $1\to 2$ & 4.885 & 14.27 & 9.578 & 11.063 & 26.462 & 17.388\\   
CW2 $1\to 2$ & {\bf 3.061} & {\bf 3.131} & {\bf 3.093} & {\bf 3.066} &
{\bf 3.14} & {\bf 3.096} \\
BIM $L_1$  $1\to 2$ & 2.915 & 3.455 & 3.165 & 3.091 & 3.514 & 3.164\\
NF   $1\to 2$ & {\bf 4.881} & {\bf 5.729} & {\bf 5.305} & {\bf 4.986}
& {\bf 8.803} &	{\bf 6.806}\\
FGSM  $1\to 2$& {\bf 14.259} & {\bf 14.41} &{\bf 14.335}&{\bf 14.413}&
{\bf 15.213}&	{\bf 14.784} \\\hline
CW2  $1\to 3$& 7.948&	11.91&	8.879&	5.663&	16.674&	8.467 \\  
BIM $L_1$  $1\to 3$& 5.195&	6.287&	5.641&	5.531&	6.43&	5.63\\
BIM $L_2$  $1\to 3$&4.403&	6.081&	5.042&	5.504&	7.732&	5.631 \\
MI  $1\to 3$& 10.406&	11.112&	10.695&	16.457&	21.854&	20.135 \\\hline
CW2  $1\to 4$&6.385&	6.87&	6.528&	6.591&	6.924&	6.743 \\\hline
CW2  $1\to 5$ &10.743&	12.989&	11.567&	9.932&	16.703&	12.298 \\
MI  $1\to 5$& 17.502&	19.526&	18.514&	34.122&	48.086&	43.28 \\\hline
CW2  $1\to 6$&7.285&	8.435&	7.86&	7.485&	12.687&	9.719  \\
BIM $L_2$  $1\to 6$& 6.557&9.689&8.123&	7.101&	9.891&	8.641 \\\hline
CW2  $1\to 7$& 4.493&	4.716&	4.575&	4.519&	4.721&	4.58 \\
BIM $L_1$  $1\to 7$& {\bf 5.039}&{\bf 7.565}&{\bf 6.002}&
{\bf  5.321}&{\bf 7.614}&{\bf 5.69}\\
BIM $L_{\infty}$  $1\to 7$&8.197&	9.124&	8.661&	13.118&	15.11&	14.855 \\
MI  $1\to 7$& 8.892&	9.457&	9.075&	14.633&	18.742&	17.377\\\hline
PW  $1\to 8$& 5.205&	14.84&	10.023&	12.526&	26.526&	17.329\\
CW2  $1\to 8$& 3.604&	4.281&	3.843&	3.722&	4.572&	3.917\\
BIM $L_1$  $1\to 8$&3.316&	4.009&	3.562&	3.637&	4.188&	3.778 \\
MI  $1\to 8$&8.569&	9.158&	8.863&	9.615&	14.493&	13.126\\\hline
CW2 $1\to 9$&11.062&	11.276&	11.069&	10.975&	12.52&	11.586 \\
BIM $L_1$  $1\to 9$& 8.636&	12.579&	10.608&	10.38&	14.186&	11.741\\\hline
CW2 $1\to 0$ &{\bf 10.099}&{\bf	11.218}&{\bf 10.644}&{\bf 11.661}
&{\bf  11.932}&{\bf 11.813} \\
BIM $L_1$ $1\to 0$&9.231&	11.994&	10.255&	12.733&	20.247&	14.552 \\
BIM $L_2$ $1\to 0$&11.299&	16.77&	14.035&	12.279&	19.563&	14.216 \\
BIM $L_{\infty}$ $1\to 0$&12.392&	15.827&	14.11&	34.519&	79.041&	39.734  \\
MI $1\to 0$&21.261&	22.607&	21.852&	43.76&	43.297&	38.818\\
\hline
\end{tabular} }
\end{center}
\end{table}

\begin{table}[t]
\caption{CNN Attack Misclassification Rates}
\label{tab:attack-misrate-mnist-cnn}
\begin{center}
{\footnotesize  
\begin{tabular}{|l|llllllllll|} \hline
   &$M_1$&$M_2$&$M_3$&$M_4$&$M_5$&$M_6$&$M_7$&$M_8$&$M_9$&$M_{10}$\\\hline
PW 38d $1\to 2$ & 1 & 0.02 & 0 & 0.02 & 0.04 & 0.27 & 0.23 & 0.27 & 0.30 & 0.49 \\   
CW2 286d $1\to 2$& 1 & 0 & 0 & 0 & 0 & 0 & 0 & 0 & 0 & 0 \\
BIM $L_1$ 425d $1\to 2$& 1 & 0 & 0 & 0 & 0 & 0 & 0 & 0 & 0 & 0\\
NF 403d  $1\to 2$& 1 & 0.82 & 0.62 & 0.92 & 0.96 & 1 & 1 & 0.95 & 1 & 1 \\
FGSM 403d $1\to 2$& 1 & 0 & 0 & 0 & 0 & 0 & 0 & 0 & 0 & 0 \\\hline
CW2 263d $1\to 3$&  1 & 0 & 0 & 0 & 0 & 0 & 0 & 0 & 0 & 0 \\  
BIM $L_1$ 455d $1\to 3$& 1 & 0 & 0 & 0 & 0.01 & 0.14 & 0.01 & 0 & 0.01 & 0.05\\
BIM $L_2$ 454d $1\to 3$& 1 & 0.01 & 0.02 & 0.08 & 0.11 & 0.23 & 0.06 & 0.02 & 0.09 & 0.12\\
MI 427d $1\to 3$& 1 & 0 & 0 & 0.44 & 0.41 & 0.62 & 0.30 & 0.03 & 0.41& 0.49 \\\hline
CW2 310d $1\to 4$ & 1 & 0 & 0 & 0 & 0 & 0 & 0 & 0 & 0 & 0\\\hline
CW2 308d $1\to 5$ & 1 & 0 & 0 & 0 & 0 & 0 & 0 & 0 & 0 & 0\\
MI 489d $1\to 5$  & 1 & 1 & 1 & 1 & 1 & 1 & 1 & 1 & 1 & 1 \\\hline
CW2 318d $1\to 6$& 1 & 0 & 0 & 0 & 0 & 0 & 0 & 0 & 0 & 0\\
BIM $L_2$ 483d $1\to 6$& 1 & 1 & 0.19 & 1 & 0.12 & 0.98 & 0.46 & 0.05& 1 & 0.28\\\hline
CW2 258d $1\to 7$& 1 & 0 & 0 & 0 & 0 & 0 & 0 & 0 & 0 & 0\\
BIM $L_1$ 460d $1\to 7$& 1 & 0.02 & 0.03 & 0.03 & 0.03 & 0.02 & 0 & 0 & 0.11 & 0.32 \\
BIM $L_{\infty}$ 491d $1\to 7$& 1 & 0 & 0 & 0 & 0 & 0 & 0 & 0 & 0 & 0\\
MI 473d $1\to 7$ & 1 & 0 & 0 & 0 & 0 & 0 & 0 & 0 & 0 & 0.30 \\\hline
PW 40d $1\to 8$& 1 & 0.04 & 0.01 & 0.15 & 0.16 & 0.55 & 0.37 & 0.41 & 0.33 & 0.48\\
CW2 258d $1\to 8$& 1 & 0 & 0 & 0 & 0 & 0 & 0 & 0 & 0 & 0 \\
BIM $L_1$ 424d $1\to 8$& 1 & 0 & 0 & 0 & 0 & 0.07 & 0 & 0 & 0 & 0 \\
MI 438d $1\to 8$& 1 & 0 & 0 & 0.55 & 0 & 0.05 & 0 & 0 & 0 & 0 \\\hline
CW2 314d $1\to 9$& 1 & 0 & 0 & 0 & 0 & 0 & 0 & 0 & 0 & 0\\
BIM $L_1$ 492d $1\to 9$& 1 & 0.55 & 0.94 & 0.98 & 0.99 & 0.99 & 0.90 & 0.74 & 0.93 & 0.96 \\\hline
CW2 286d $1\to 0$ & 1 & 0 & 0 & 0 & 0 & 0 & 0 & 0 & 0 & 0\\
BIM $L_1$ 496d $1\to 0$& 1 & 0.91 & 0.62 & 0.83 & 0.76 & 0.89 & 0.79 &0.68 & 0.92 & 0.91 \\
BIM $L_2$ 495d $1\to 0$& 1 & 0.91 & 0.30 & 0.70 & 0.55 & 0.85 & 0.63 & 0.43 & 0.76 & 0.76\\
BIM $L_{\infty}$ 510d $1\to 0$& 1 & 1 & 0.98 & 0.98 & 0.96 & 0.98 & 0.87 & 0.52 & 0.59 & 0.52 \\
MI 462d $1\to 0$& 1 & 1 & 0.95 & 1 & 0.87 & 0.85 & 0.82 & 0.62 & 0.63& 0.67 \\
\hline
\end{tabular} }
\end{center}
\end{table}

\begin{figure}[t]
\centering
\subfloat[FGSM $1\to 2$]{\includegraphics[width=0.25\textwidth]{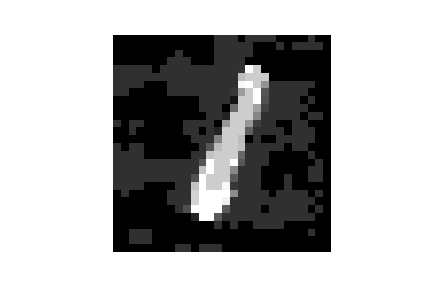}}
\centering
\subfloat[MI $1\to 3$]{\includegraphics[width=0.25\textwidth]{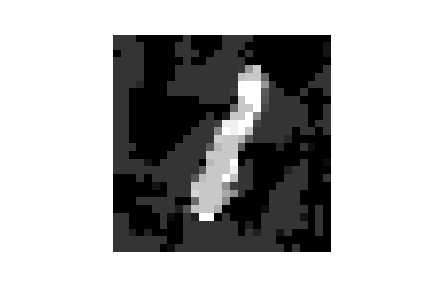}}
\centering
\subfloat[CW2 $1\to 4$]{\includegraphics[width=0.25\textwidth]{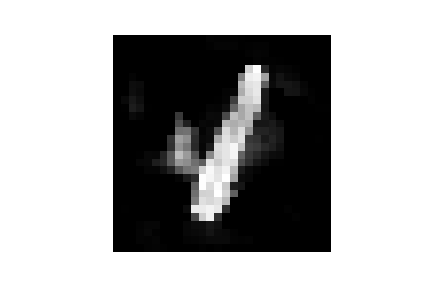}}
\\
\centering
\subfloat[CW2 $1\to 5$]{\includegraphics[width=0.25\textwidth]{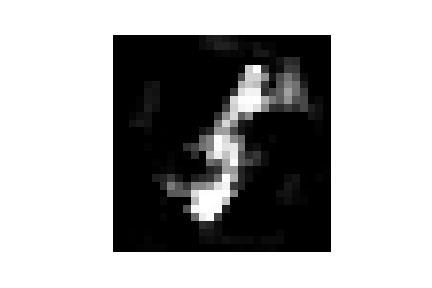}}
\centering
\subfloat[BIM $L_2$ $1\to 6$]{\includegraphics[width=0.25\textwidth]{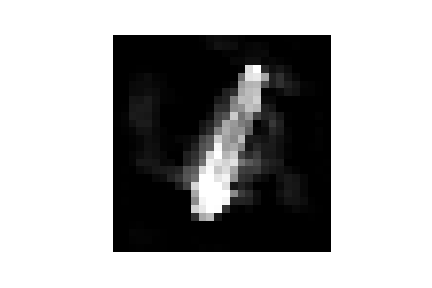}}
\centering
\subfloat[MI $1\to 7$]{\includegraphics[width=0.25\textwidth]{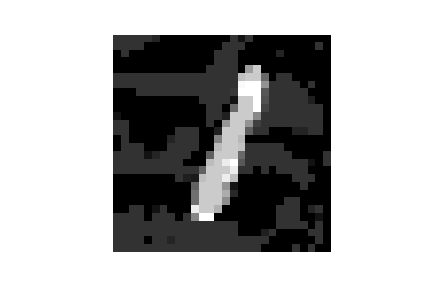}}
\\
\centering
\subfloat[PW $1\to 8$]{\includegraphics[width=0.25\textwidth]{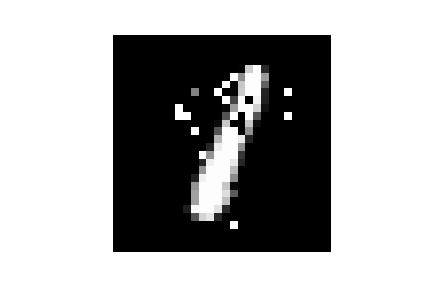}}
\centering
\subfloat[BIM $L_{\infty}$ $1\to 9$]{\includegraphics[width=0.25\textwidth]{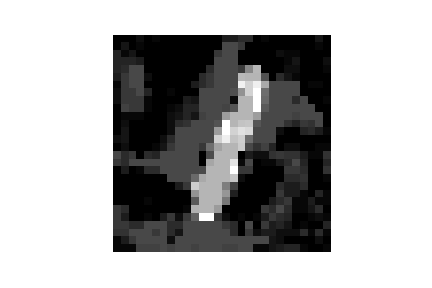}}
\centering
\subfloat[BIM $L_1$ $1\to 0$]{\includegraphics[width=0.25\textwidth]{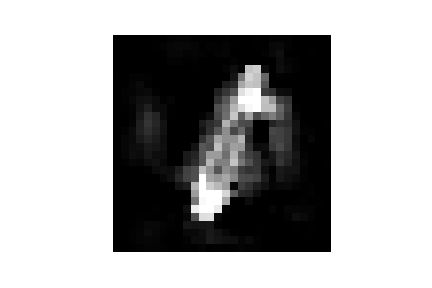}}
\caption{Adversarial Examples Generated by Attack Algorithms}
\label{fig:attack-images}
\end{figure}

\begin{figure}[t]
\centering
\subfloat[FGSM 375d $1\to 2$]{\includegraphics[width=0.25\textwidth]{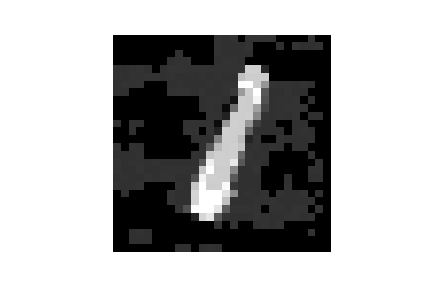}}
\centering
\subfloat[MI 250d $1\to 3$]{\includegraphics[width=0.25\textwidth]{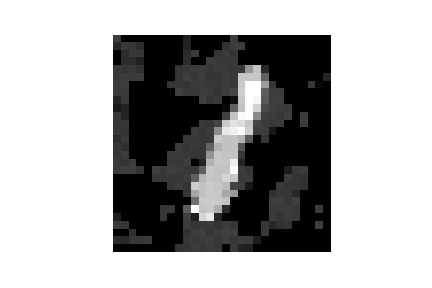}}
\centering
\subfloat[CW2 180d $1\to 4$]{\includegraphics[width=0.25\textwidth]{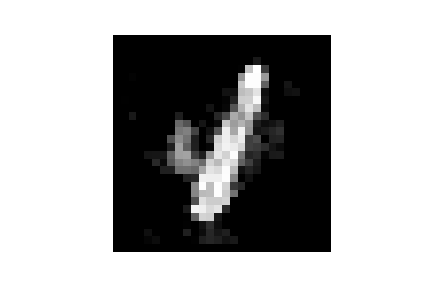}}
\\
\centering
\subfloat[CW2 175d $1\to 5$]{\includegraphics[width=0.25\textwidth]{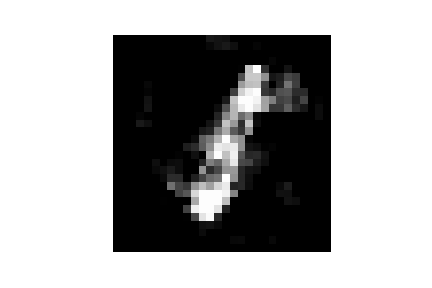}}
\centering
\subfloat[BIM $L_2$ 220d $1\to 6$]{\includegraphics[width=0.25\textwidth]{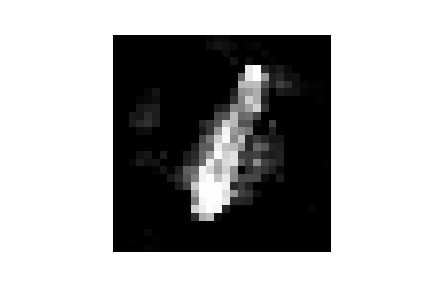}}
\centering
\subfloat[MI 300d $1\to 7$]{\includegraphics[width=0.25\textwidth]{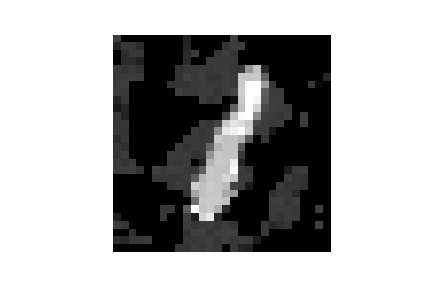}}
\\
\centering
\subfloat[PW 35d $1\to 8$]{\includegraphics[width=0.25\textwidth]{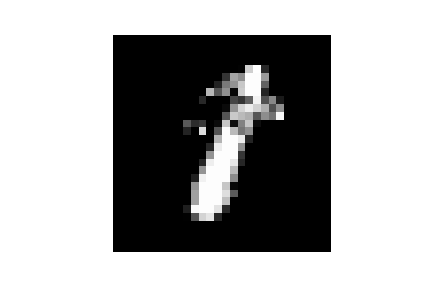}}
\centering
\subfloat[BIM $L_{\infty}$ 110d $1\to 9$]{\includegraphics[width=0.25\textwidth]{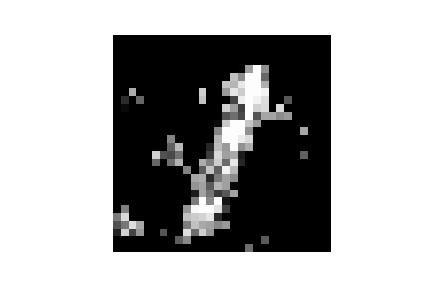}}
\centering
\subfloat[BIM $L_1$ 120d $1\to 0$]{\includegraphics[width=0.25\textwidth]{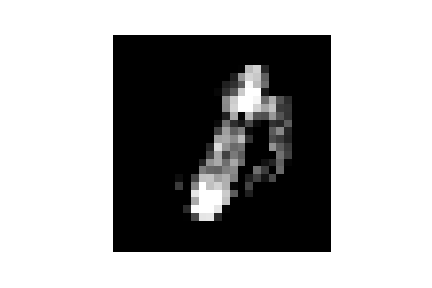}}
\caption{Adversarial Examples Generated by Sampling in
  Hyper-Rectangles $R_k(t)$}
\label{fig:delta-ball-images}
\end{figure}

\paragraph{Uncertainty Region Construction}
Assume $M_1$ is the model under attack. 
For a given attack algorithm and a object class $t$, $t\neq c$ where
$c$ is the true object class of $W$, we
combine the adversarial examples $W^a$ from both the targeted attack
and the untargeted attack, s.t. $M_1(W^a)=t$. We then construct the
subspace spanned by $W^a$. This step requires an attack algorithm to
generate sufficient amount of perturbed images $W^a$, at least 80-100
images, given a clean
image. Although there are 42 attacks
algorithms in Foolbox, most of them cannot satisfy this requirement,
including several famous attacks -- DeepFool attack 
\cite{attack-deepfool}, L-BFGS attack \cite{attack-LBFGS}, PGD attack 
\cite{attack-PGD} on MNIST. Furthermore we only consider adversarial examples
$W^a$ in $B(\delta,W)$ with a small $\delta$. Some attacks generate
large perturbations that are barely recognizable, such as Spatial
Transform Attack \cite{attack-spatial}, 
Additive Gaussian Noise Attack \cite{foolbox-2017}, and
Additive Uniform Noise Attack \cite{foolbox-2017}. They are also
excluded from the experiments. We only use the attack algorithms
listed in Sec.~\ref{sec:related} in the experiments. 

Let $I_k(t):=\left\{W^a_k=(W^a_{k,i}):~M_1(W^a_k)=t,~
t\neq c \right\}$. $I_k(t)$ is the set of adversarial images 
misclassified to class $t$ by attack algorithm $k$. If $\exists~
W^a_{k,i}\neq W_i$, i.e., attack $k$ adds perturbation to the $i$th
pixel, we compute the interval size of the $i$th pixel as
$s^{t,k}_i = max_{I_k(t)}(W^a_{k,i}) - min_{I_k(t)}(W^a_{k,i})$. Assume $m$
pixels are perturbed by attack $k$. Then the intervals 
are ranked by interval size as $s^{t,k}_{(1)}\geq s^{t,k}_{(2)} \geq \cdots
s^{t,k}_{(m)}$. We construct a hyper-rectangle $R_k(t)$ using $b$ largest
intervals with $b \leq m$ as
$$ R_k(t) ~=~ [min_{I_k(t)}(W^a_{k,(1)}),~max_{I_k(t)}(W^a_{k,(1)})] \otimes \cdots  
\otimes [min_{I_k(t)}(W^a_{k,(b)}),~max_{I_k(t)}(W^a_{k,(b)})].$$
$R_k(t)$ is the subspace based on the adversarial examples generated by
attack $k$ and misclassified to class $t$. We choose the
number of intervals $b$ that the remaining interval sizes are very
small and the perturbations added can be considered as approximately constant.

\subsection{MNIST CNN Experiment}
\label{sec:mnist}

\begin{table}[b]
\centering
\caption{MLP Mis-classification Rates on Clean MNIST Test Images}
\label{tab:MLP-baseline}
\begin{center}
  {\footnotesize
\begin{tabular}{|lllll|}
\hline
$M_1$ & $M_2$ & $M_3$ & $M_4$ & $M_5$ \\\hline 
0.0177 & 0.0178 & 0.0183 &  0.0177 &  0.0176 \\
\hline 
\end{tabular} }
\end{center}
\end{table}

\begin{table}[b]
\caption{MLP Misclassification Rates in Hyper-Rectangles}
\label{tab:hyper-rec-misrate-MLP}
\begin{center}
{\footnotesize
\begin{tabular}{|l|l|lllll|} \hline
  &$s_{(i)}$ &$M_1$&$M_2$&$M_3$&$M_4$&$M_5$\\\hline
CW2 230d $5\to 6$& 0.005 & 0.94 & 0 & 0 & 0 & 0 \\
CW2 440d $7\to 2$& 0.01 & 0.82 & 0 & 0.78 & 0 & 0 \\
\hline
\end{tabular} }
\end{center}
\end{table}

\begin{table}[b]
\caption{MLP Attack Misclassification Rates}
\label{tab:cw2-misrate-MLP}
\begin{center}
{\footnotesize 
\begin{tabular}{|l|lllll|} \hline
   &$M_1$&$M_2$&$M_3$&$M_4$&$M_5$\\\hline
CW2 380d $5\to 6$& 1 & 0 & 0 & 0 & 0  \\
CW2 491d $7\to 2$& 1 & 0 & 0.28 & 0 & 0  \\
\hline
\end{tabular} }
\end{center}
\end{table}

\begin{figure}[t]
\centering
\subfloat[Clean $5$]{\includegraphics[width=0.25\textwidth]{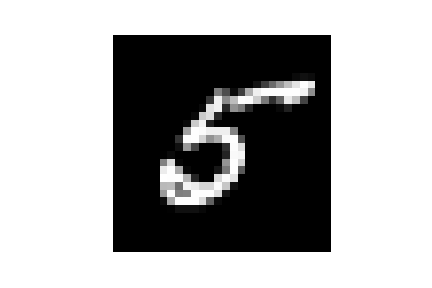}}
\centering
\subfloat[CW2 $5\to 6$]{\includegraphics[width=0.25\textwidth]{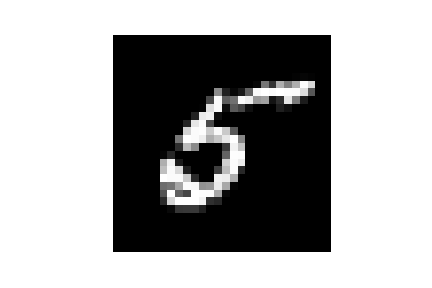}}
\centering
\subfloat[Sampled $5\to 6$]{\includegraphics[width=0.25\textwidth]{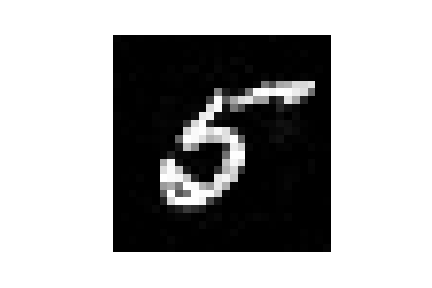}}
\\
\centering
\subfloat[Clean $7$]{\includegraphics[width=0.25\textwidth]{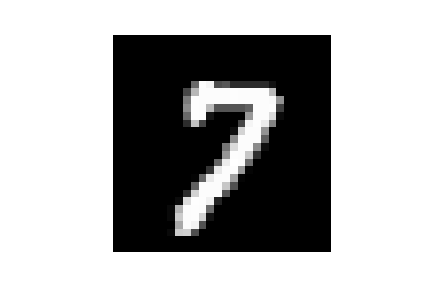}}
\centering
\subfloat[CW2 $7\to 2$]{\includegraphics[width=0.25\textwidth]{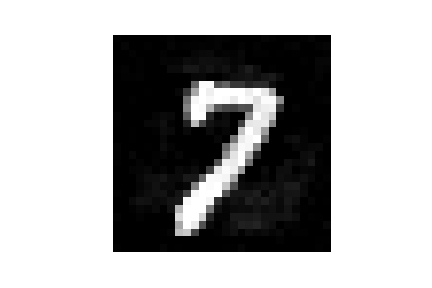}}
\centering
\subfloat[Sampled $7\to 2$]{\includegraphics[width=0.25\textwidth]{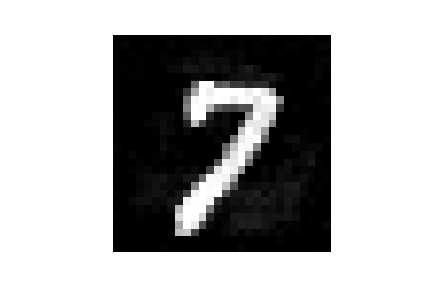}}
\caption{Images for MLP experiment}
\label{fig:MLP-images}
\end{figure}

Here we conduct an experiment with the task to
classify MNIST dataset of 10 handwritten digits \cite{mnist}. MNIST has
60,000 training 
images and 10,000 test images. Each image has 28x28 gray-scale
pixels. Our model is the PyTorch implementation of LeNet
\cite{lenet1998}, which has
two convolutional layers. The model structure can be found in \cite{pytorch-lenet-fgsm,pytorch-lenet-tutorial}. The pixels are rescaled to
$[0,1]$ in PyTorch implementation. $W$ is a vectorized MNIST
image. We have $W \in [0,1]^{784}$. Table~\ref{tab:baseline} shows the
accuracy of 10 trained LeNet models on the MNIST test data using
different initial  seeds.  $M_1$ to
$M_{10}$ have similar performance.   

Intuitively, the ten handwritten digits have distinct features that
facilitate the classification task. Hence LeNet can achieve nearly 99\%
accuracy. We visualize the digits 
using t-Distributed Stochastic Neighbor Embedding (t-SNE) technique
\cite{tSNE-dimreduction-2008}, a nonlinear dimension reduction
technique. t-SNE uses t distribution to 
calculate the similarity between two points and can capture the local
properties of high dimensional data that lie on lower dimensional
manifold. Figure~\ref{fig:2D} provides a two dimensional projection of
the ten digits, based on 2000 sampled images. The digits form tight clusters.

In the experiment, we choose a clean image $W$, and generate
adversarial examples using the attack algorithms listed in
Section~\ref{sec:related}. We then construct the 
hyper-rectangles $R_k(t)$ that 
contain the adversarial examples. We  have studied many test images and
training images, and  have obtained similar results. Due to the limited
space, here we show the 
results for a digit 1 from the test data, shown in
Figure~\ref{fig-1}. We run the attacks against $M_1$. 
Table~\ref{tab:hyper-rec-misrate} shows the following information.
\begin{enumerate}
\item The number of intervals used to construct the
  hyper-rectangles. For example, CW2 280d $1\to 2$ means $R_{CW2}(2)$
  is spanned by the largest 280 intervals.
  
\item The smallest interval size in $R_k(t)$, shown in column
$s_{(i)}$. For PW attack, we use  [0,1] for the selected
pixels, since the measured interval sizes are all close to 1. For all other
attacks, the interval size is measured from the added perturbations.
\item We sample 1000 images from each hyper-rectangle $R_k(t)$, and report the
  misclassification rates by $M_1$ to $M_{10}$. 
\end{enumerate}
The left three columns in Table~\ref{tab:L2} show the minimum amount
of perturbations ($min(||W^a-W||_2)$), the maximum amount of
perturbations ($max(||W^a-W||_2)$), and the average amount of
perturbations ($mean(||W^a-W||_2)$) of the 1000 sampled images in
each hyper-rectangle $R_k(t)$. The right three columns in
Table~\ref{tab:L2} show the same information for the adversarial
examples generated by the attacks. Figure~\ref{fig:attack-images}
shows the adversarial examples generated by the 
attacks. Figure~\ref{fig:delta-ball-images} shows the adversarial  
images generated by sampling in the hyper-rectangles $R_k(t)$. 

As shown in Table~\ref{tab:hyper-rec-misrate}, the column of $s_{(i)}$
has the maximum value 0.036. This 
translates to 12 consecutive integer values on the original 0, 1, ...,
255 scale. They are very similar light levels, and can be considered as
approximately constant. If we
add more dimensions to $R_k(t)$, the additional dimensions can be considered as
moving the additional pixels to different values. Adding more dimensions do not
change the shape and size of $R_k(t)$. Instead that 
moves the hyper-rectangle to a different location, increasing the
amount of perturbation and away from the clean image $W$. 
The hyper-rectangles $R_k(t)$ in Table~\ref{tab:hyper-rec-misrate}
used far fewer pixels than the original attacks. From
Table~\ref{tab:L2}, we see that leads to smaller perturbations
to create adversarial examples in $R_k(t)$. There are more
such hyper-rectangles with the same shape and size, as we add more
pixels identified by the attacks. 
Adding more pixels does not necessarily increase the misclassification
rates by all DNNs. For Carlini \& Wagner $L_2$ attack
and FGSM, eventually the hyper-rectangle is moved to a place where $M_1$
misclassification rate is close to 100\% and all other models, $M_2$ to
$M_{10}$, see near 0\% misclassification rate. This is the effect of
the optimization approach attacking $M_1$.

As highlighted in Table~\ref{tab:hyper-rec-misrate},
we observe three types of $R_k(t)$:
(1) the target DNN misclassifies most of the adversarial
examples while  there exists another DNN which correctly classifies the
adversarial examples; (2) the target model correctly classifies the
adversarial examples while another DNN misclassifies most of the
adversarial examples; (3) the transferable adversarial regions where
all DNNs misclassify a significant proportion of the adversarial
examples. The highlighted rows in Table 3 shows this phenomenon occurs
to attacks adding both small and large perturbations.
The first two types of $R_k(t)$ belong to DNN uncertainty
regions. The existence of DNN uncertainty regions 
shows transferability of adversarial examples is not universal, contrary to
\cite{attack-LBFGS,explain-adexample-2015,transferable-adexample-space-2017}.
A better understanding of DNN classification boundary, its 
uncertainty regions, and transferable adversarial regions, is crucial
to improve the robustness of DNN.  
 
Table~\ref{tab:attack-misrate-mnist-cnn} shows the attacks'
misclassification rates on the 10 LeNet models, and the number of perturbed
pixels. Adversarial examples by Carlini \& Wagner $L_2$ attack are
correctly classified by the other 9 models. 
Compared to Carlini \& Wagner $L_2$ attack, BIM $L_1$, $L_2$,
$L_{\infty}$, NewtonFool, Moment Iterative attacks perturbed a lot more
pixels, between 400 to 600, to generate adversarial examples that cause
misclassification from multiple trained DNN models.  
Notice the hyper-rectangles by our approach, having smaller
number of perturbed pixels than the original attacks, can attack
several DNNs simultaneously. 

\begin{table}[t]
\centering
\caption{MobileNet Mis-classification Rates on Clean CIFAR10 Test Images}
\label{tab:MobileNet-baseline}
\begin{center}
  {\footnotesize
\begin{tabular}{|lllll|}
\hline
$M_1$ & $M_2$ & $M_3$ & $M_4$ & $M_5$ \\\hline 
0.0767 & 0.0728 & 0.0734 & 0.0727 &  0.0744 \\
\hline 
\end{tabular} }
\end{center}
\end{table}

\begin{table}[t]
\caption{MobileNet Attack Misclassification Rates}
\label{tab:attack-misrate-MobileNet}
\begin{center}
{\footnotesize
\begin{tabular}{|l|lllll|} \hline
  &$M_1$&$M_2$&$M_3$&$M_4$&$M_5$\\\hline
BIM $L_2$ 3017d airplane$\to$deer & 1 & 0 & 0.88 & 0 & 0\\
\hline
\end{tabular} }
\end{center}
\end{table}

\begin{figure}[t]
\centering
\subfloat[Clean airplane]{\includegraphics[width=0.25\textwidth]{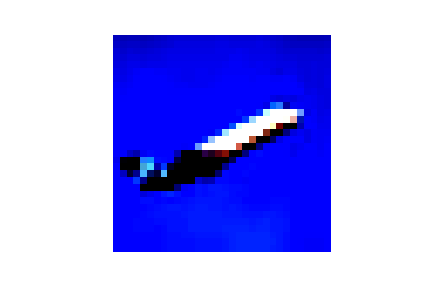}}
\centering
\subfloat[BIM $L_2$ airplane$\to$deer]{\includegraphics[width=0.25\textwidth]{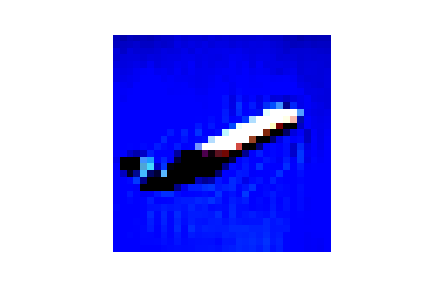}}
\centering
\subfloat[Sampled airplane$\to$deer]{\includegraphics[width=0.25\textwidth]{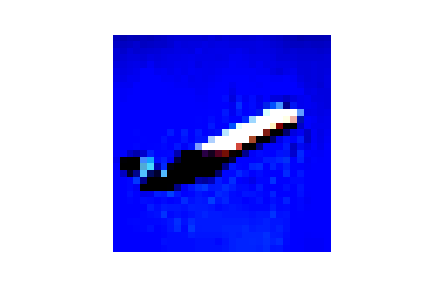}}
\caption{Images for MobileNet experiment}
\label{fig:MobileNet-images}
\end{figure}

\begin{figure}[t]
\centering
\centering
\includegraphics[width=3.5in]{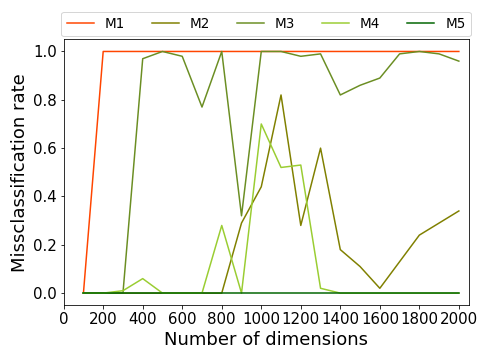}
\caption{MobileNet misclassification rates in hyper-rectangles}
\label{fig:MobileNet-plotresults}
\end{figure}

\subsection{MNIST MLP Experiment}
\label{sec:MLP}
Here we conduct experiment with a MLP trained on
MNIST. It is a fully connected network with 3 layers, 3x512 hidden neurons and ReLU
activation. We vary the initial seeds and train 5
MLPs. Table~\ref{tab:MLP-baseline} shows the MLP misclassification
rates on the clean MNIST test data. They have more consistent
performance compared to the 10 CNN
models. In the interest of space, here we
show two examples, a digit 5 and a digit 7, under Carlini \& Wagner $L_2$ attack. 
Table~\ref{tab:hyper-rec-misrate-MLP} shows the 5 
MLPs' misclassification rates in the
hyper-rectangles. Table~\ref{tab:cw2-misrate-MLP} shows the attacks'
misclassification rates. Figure~\ref{fig:MLP-images} shows the
clean images, adversarial examples generated by Carlini \& Wagner
$L_2$ attack, and adversarial examples generated by sampling from hyper-rectangles. 
The misclassification rates suggest the hyper-rectangles and the
adversarial images generated by Carlini \& Wagner $L_2$ attack lie in
DNN's uncertainty regions. Again Carlini \& Wagner $L_2$ attack has
great success with the target model $M_1$ but can be blocked by the
other four MLPs. 

\begin{table}[t]
\caption{$L_2$ Distance for MLP and MobileNet Experiments}
\label{tab:L2-mlp-cifar10}
\begin{center}
{\footnotesize  
\begin{tabular}{|l|lll|lll|} \hline
  &$L_2^{min}$ & $L_2^{max}$  &  $\bar{L}_2$ 
  &$L_2^{min}$ Attack& $L_2^{max}$ Attack & $\bar{L}_2$ Attack\\\hline
CW2 MLP 230d $5\to6$  & 10.47 & 10.72 & 10.61 &  11.06 &  11.51 & 11.26\\
CW2 MLP 440d $7\to2$   & 11.18 & 11.41 & 11.30  &  11.51 & 11.89 & 11.68\\ \hline
BIM $L_2$ 2000d airplane$\to$deer & 64.49 & 67.05 & 65.88 & 225 & 288.32 & 256.01  \\\hline
\end{tabular} }
\end{center}
\end{table}

\subsection{CIFAR10 MobileNet Experiment}
We re-train the 
MobileNet~\cite{mobilenetv2-2018} on CIFAR10 \cite{cifar10}. CIFAR10
has 60,000 32x32 color images in 10 classes, with 50,000 as training images
and 10,000 as test images. A vectorized CIFAR10 image is in
$[0,1]^{3072}$. The dimensionality of a CIFAR10 image is almost 4
times of a MNIST image.
MobileNet has an initial convolution layers followed
by 19 residual bottleneck layers. \cite{mobilenetv2-2018} shows the
network structure. MobileNet has a more complex model structure
aiming to reduce memory usage.  
The misclassification rates of five re-trained MobileNet models on the
clean test data by varying
initial seeds are in Table~\ref{tab:MobileNet-baseline}. 
For the interest of space, here we show an example with an airplane image
under BIM $L_2$ attack. The attack success on the five MobileNet
models are in Table~\ref{tab:attack-misrate-MobileNet}. Note BIM $L_2$
attack perturbed 3071 dimensions and left 1 dimension untouched. 
The images are shown in Figure~\ref{fig:MobileNet-images}. 

Figure~\ref{fig:MobileNet-plotresults} shows the misclassification
rates as we increase the dimensions of the hyper-rectangle. 
The largest interval size is 0.2 and the 2000th largest interval size
is 0.017. $M1$ misclassifies all the sampled images starting from 
around 200 perturbed dimensions. $M5$ correctly classifies all the
sampled images. We see $M2$ and $M4$ misclassification rates increase
as more effective dimensions are included, then decrease as we
include additional irrelevant dimensions. The 2000-dimensional
hyper-rectangle is a MobileNet uncertainty region. 
As noted in \cite{explain-adexample-2015}, the direction of
adversarial perturbation is important. Adversarial examples cannot be
generated by randomly sampling in 3072 dimensional $\delta-$
neighborhood $B(\delta,W)$. The lower dimensional
hyper-rectangles $R_k(t)$ containing infinitely many 
adversarial examples are discovered through
optimization approach, i.e., the attack algorithms. 
Table~\ref{tab:L2-mlp-cifar10} shows that the sampled adversarial
images from the hyper-rectangle have much smaller perturbations than
the original attack on CIFAR10.  

\paragraph{Strategy for Robust Classification} Our experiments
confirm that transferability of adversarial examples is not
universal. Notice all the trained DNNs achieve the same level of
accuracy over the clean test images, but their performance vary a lot
over the hyper-rectangles.  Although in some hyper-rectangles, all the
trained models misclassify a large portions of the sampled
adversarial images, in 
uncertainty regions, there exist at least one DNN that can correctly
classify all the adversarial images. This naturally leads to a
strategy to make robust decision. If at least one DNN assigns
a label that is different from another DNN, the image triggers
an alert and requires additional screening, either involving a human
operator or alternative classifiers. This strategy will significantly
improve the accuracy over the adversarial examples in DNN uncertainty
regions, but won't solve the problem for transferable adversarial
examples. Although an ensemble can mitigate many attacks, based on
Tables~\ref{tab:attack-misrate-mnist-cnn}, \ref{tab:cw2-misrate-MLP}, \ref{tab:attack-misrate-MobileNet},
unfortunately an ensemble of DNNs may not achieve the 
desirable accuracy over uncertainty regions. There is no
guarantee about the number of DNNs that can make correct
decision over each uncertainty region.
We also need to understand how to measure the size of DNN
uncertainty regions vs. DNN transferable adversarial regions. We leave
it to the future work.    

\section{Generalization Error and Adversarial Examples}
\label{sec:theory}

The classification accuracy on test data is often used to measure a
classifier's performance. However, in \cite{ge-definition-bengio-2003}, the
authors argued the test data accuracy is not the most appropriate performance
measure, because the variability due to the randomness of the
training data needs to be 
taken into consideration, besides those due to the test data. 
Let $Z = (W,Y)$ denote a sample, where $W \in [0,1]^h$ is the $h-$dimensional
vectorized image features and $Y \in \{1, \cdots, c\}$ is the true
object class. $Z$ is generated independently and identically from a
distribution $F$ over $[0,1]^h$. We denote a training 
dataset with $n$ sample points by $D_n=(Z_1, \cdots, Z_n)$.
\cite{ge-definition-bengio-2003} defined
generalization error as $E(loss_M(D_n,Z_{n+1}))$, where 
$Z_{n+1}$ is a test sample, and $loss_M()$ is the loss of applying a 
classifier $M$ trained on $D_n$ to $Z_{n+1}$. If $loss_M()$ is a 0-1
loss, the generalization error is 
defined as the error probability $P(M(W)\neq Y)$  in
\cite{ge-svm-mnist-2005}.   
Another definition of generalization error involves the empirical
error on the training data. 
Let $\hat{loss}_{M}(D_n) = \frac{1}{n} \sum_1^n loss(Z_i)$ be the empirical
risk estimated from the training data $D_n$.
\cite{ge-rethink-2017} defined generalization error as
$E(loss_{M}(D_n,Z_{n+1})) - \hat{loss}_{M}(D_n)$, which is widely used
in many recent papers to establish DNN theoretical guarantees. 

There is an extensive literature on theoretical generalization error bound, for different type of classifiers including DNN. 
\cite{ge-definition-bengio-2003} suggested the
generalization error can be estimated through cross
validation. Experiments were conducted on least square linear
regression, regression tree, classification tree and nearest neighbor
classifier. They built normal
confidence interval or confidence intervals based on Student's t
distribution for generalization error.   
Subsequently, \cite{ge-svm-mnist-2005} obtained generalization error
bound based on the disagreement probability between classifiers, and
computed the generalization error bound of SVM on MNIST data.   
Generalization error bound for DNN is proven to be
$O(\frac{c(depth,width)}{\sqrt{n}})$, where $c(depth,width)$ refers to
a constant based on the width and depth of a DNN model, e.g., in 
\cite{ge-bound-1,ge-bound-2,ge-bound-3,ge-bound-4,ge-bound-5,ge-bound-6,ge-bound-7,ge-bound-8}.  

We observe there is a discrepancy between the theoretically proven
generalization error bound for DNN and the existence of adversarial
examples. Following the theory, the error on test data should
decrease to 0 at a rate proportional to $n^{-1/2}$ where $n$ is the
training sample size. But given a clean
image, we show there exists infinitely many adversarial images in
$B(\delta,W)$ for different network structures and
datasets. Adversarial examples exist for large DNN models trained on
ImageNet with millions of training data, where the theoretical
asymptotic behavior of DNN should already kick in. Here we offer an
explanation for why this happens. 
Let $L_r$ be a $r-$dimensional region in $[0,1]^h$ with $r<h$. Let
$\mathcal{L} = \cup_{i=1}^{\infty} L_{r_i}$ 
be the union of countably infinite non-overlapping lower dimensional
regions $L_{r_i}$ in  $[0,1]^h$ with all $r_i<h$.

\begin{theorem}
Let $M_1$ and $M_2$ be two DNN
models trained on $D_n$. Assume $\forall~ W \in [0,1]^h-\mathcal{L}$,
$M_1(W)= M_2(W)$. And assume $\exists~ W \in \mathcal{L}$, s.t.
$M_1(W) \neq  M_2(W)$. We have
$$E(loss_{M_1}(D_n,Z_{n+1})) = E(loss_{M_2}(D_n,Z_{n+1})).$$
\end{theorem}

\begin{proof}
For any continuous distribution $F$ on $[0,1]^h$,
$F(\mathcal{L})=0$, i.e., the lower dimensional region $\mathcal{L}$
has 0 probability mass. For two functions that differ only on 0
probability region, we have
$$E(loss_{M_1}(D_n,Z_{n+1})) = E(loss_{M_2}(D_n,Z_{n+1})).$$
\end{proof}

\begin{corollary}
There exist $M_1$ and $M_2$ s.t.    
$$E(loss_{M_1}(D_n,Z_{n+1})) - \hat{loss}_{M_1}(D_n) =
E(loss_{M_2}(D_n,Z_{n+1})) - \hat{loss}_{M_2}(D_n).$$  
\end{corollary}

\begin{proof}
Corollary 1 in \cite{ge-rethink-2017} states there exists neural networks
with  ReLU activation, depth $g$, width $O(n/g)$ and weights
$O(n+h)$, that can fit exactly any function on $D_n$ in $h-$dimensional
space. Assume $M_1$ and $M_2$ are such models trained on
$D_n$. $\hat{loss}_{M_1}(D_n)=\hat{loss}_{M_2}(D_n)=0$. 
By only changing initial seeds during training, $M_1$ and $M_2$
satisfy the conditions in Theorem 1.  
Consequently we have 
$$E(loss_{M_1}(D_n,Z_{n+1})) - \hat{loss}_{M_1}(D_n) =
E(loss_{M_2}(D_n,Z_{n+1})) - \hat{loss}_{M_2}(D_n).$$
\end{proof}  

\paragraph{Remark} The volume of a $h-$dimensional $B(\delta,W)$ is
$|B(\delta,W)|=\frac{\pi^{h/2}}{\Gamma(1+h/2)} \delta^h$. The volume
of the feature space $[0,1]^h$ is 1. Hence for a fixed
$\delta$ and $h$, there are only finite
number of non-overlapping $\delta$ balls in the feature space
$[0,1]^h$. As $h \longrightarrow \infty$, we have $|B(\delta,W)|
\longrightarrow 0$. Hence the feature space for higher resolution color
images can contain increasingly more clean images. The existing attack
algorithms collectively identify finite number of lower dimensional
regions in $B(\delta,W)$ for each clean image. Theorem 1 means
$E(loss_{M_1}(D_n,Z_{n+1}))$ cannot tell the difference between a
trained classifier that assign correct labels to all sample points in
$B(\delta,W)$ and a different classifier that assign wrong labels only
to sample points in 
finite or countably infinite lower dimensional bounded regions. Hence
the generalization error definition and the subsequently proven large
sample property of DNN based 
on generalization error cannot adequately describe the phenomenon of
adversarial examples. We need new theory to measure DNN robustness.

\section{Conclusion}
\label{sec:conclusion}

Limitation of our work is that we rely on the existing attack
algorithms to identify these hyper-rectangles. Also our approach works
with low resolution  images. Our conjecture
is for high resolution images, adversarial examples lie in bounded
regions on lower dimensional manifold. Again we leave it to the future
work to capture the DNN classification boundary in much higher
dimensional feature space. 

\begin{figure}[h!]
\centering
\includegraphics[width=2.5in]{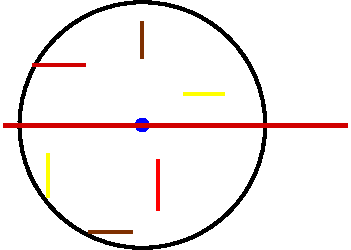}
\caption{Conceptual Plot of DNN Classification Boundary in $B(\delta)$}
\label{fig:concept}       
\end{figure}

We use Figure~\ref{fig:concept} to show a
conceptual plot of DNN classification boundary in the neighborhood $B(\delta,W)$. For the
digit 1, let $\delta=6$.  The blue dot
in the center is the clean image. 
Inside the black circle, the image
samples can be correctly 
recognized by people as 1, but there exists adversarial examples causing
misclassification errors. The hyper-rectangles are lower dimensional
small ``cracks'' inside the circle.  
There are three types of such ``cracks'', illustrated
using three different colors.
\begin{enumerate}
\item $W^a$s are misclassified by $M_1$ but can be correctly
  classified by some other model $M_j$;
\item $W^a$s are correctly classified by $M_1$ but
  misclassified by some $M_j$;
\item $W^a$s are misclassified by all the models, $M_1$ to $M_{10}$. 
\end{enumerate}
Type 1 and 2 hyper-rectangles belong to DNN uncertainty regions. Type
3 hyper-rectangles are the transferable adversarial regions which are
more difficult to handle.    
We conclude that the adversarial examples stem from a
structure problem of DNN. DNN's classification boundary is unlike that of any 
other classifier. Current defense strategies  do not address this
structural problem. We also need new theory to describe the phenomenon
of adversarial examples and measure the robustness of DNN.  


%
\section*{Acknowledgments}
This work is partially funded by ARO W911NF-17-1-0356, Northrop
Grumman Corp., and supported by SAMSI GDRR program.




\end{document}